\documentclass[acmtog]{acmart}
\acmSubmissionID{425}

\usepackage{booktabs} %
\usepackage{amsmath}
\usepackage{cleveref}
\usepackage{subcaption}
\usepackage{xcolor}
\usepackage[export]{adjustbox}

\usepackage[ruled]{algorithm2e} %

\SetAlFnt{\small}
\SetAlCapFnt{\small}
\SetAlCapNameFnt{\small}
\SetAlCapHSkip{0pt}
\IncMargin{-\parindent}

\DeclareMathOperator*{\argmin}{arg\,min}
\DeclareMathOperator{\Split}{split}

\DeclareMathOperator{\CostMatch}{C_\textrm{match}}
\DeclareMathOperator{\CostLength}{C_\textrm{len}}
\DeclareMathOperator{\CostSwap}{C_\textrm{swap}}
\DeclareMathOperator{\CostViseme}{C_\textrm{vis}}
\DeclareMathOperator{\CostInsert}{C_\textrm{insert}}
\DeclareMathOperator{\CostDelete}{C_\textrm{delete}}

\DeclareMathOperator{\EditOp}{\mathcal{W}}
\DeclareMathOperator{\PhoneSeq}{V}
\DeclareMathOperator{\SomePhonemeSubseq}{\PhoneSeq_\star}
\DeclareMathOperator{\Phone}{v}
\DeclareMathOperator{\QuerySeq}{W}
\DeclareMathOperator{\Query}{w}
\DeclareMathOperator{\BackgroundSeq}{\mathcal{B}}

\begin{document}
\title{Text-based Editing of Talking-head Video}

\author{Ohad Fried}
\affiliation{\institution{Stanford University}}
\author{Ayush Tewari}
\affiliation{\institution{Max Planck Institute for Informatics}}
\author{Michael Zollh\"{o}fer}
\affiliation{\institution{Stanford University}}
\author{Adam Finkelstein}
\affiliation{\institution{Princeton University}}
\author{Eli Shechtman}
\affiliation{\institution{Adobe}}
\author{Dan B Goldman}
\affiliation{}
\author{Kyle Genova}
\affiliation{\institution{Princeton University}}
\author{Zeyu Jin}
\affiliation{\institution{Adobe}}
\author{Christian Theobalt}
\affiliation{\institution{Max Planck Institute for Informatics}}
\author{Maneesh Agrawala}
\affiliation{\institution{Stanford University}}

\renewcommand\shortauthors{Fried, O. et al}

\newcommand*{\ShowNotes}{}
\newcommand{\ignorethis}[1]{}
\newcommand{\redund}[1]{#1}

\newcommand{\etal       }     {{et~al.}}
\newcommand{\apriori    }     {\textit{a~priori}}
\newcommand{\aposteriori}     {\textit{a~posteriori}}
\newcommand{\perse      }     {\textit{per~se}}
\newcommand{\eg         }     {{e.g.~}}
\newcommand{\Eg         }     {{E.g.~}}
\newcommand{\ie         }     {{i.e.~}}
\newcommand{\naive      }     {{na\"{\i}ve}}

\newcommand{\Identity   }     {\mat{I}}
\newcommand{\Zero       }     {\mathbf{0}}
\newcommand{\Reals      }     {{\textrm{I\kern-0.18em R}}}
\newcommand{\isdefined  }     {\mbox{\hspace{0.5ex}:=\hspace{0.5ex}}}
\newcommand{\texthalf   }     {\ensuremath{\textstyle\frac{1}{2}}}
\newcommand{\half       }     {\ensuremath{\frac{1}{2}}}
\newcommand{\third      }     {\ensuremath{\frac{1}{3}}}
\newcommand{\fourth     }     {\ensuremath{\frac{1}{4}}}

\newcommand{\degree} {\ensuremath{^{\circ}}}

\newcommand{\mat        } [1] {{\text{\boldmath $\mathbit{#1}$}}}
\newcommand{\Approx     } [1] {\widetilde{#1}}
\newcommand{\change     } [1] {\mbox{{\footnotesize $\Delta$} \kern-3pt}#1}

\newcommand{\Order      } [1] {O(#1)}
\newcommand{\set        } [1] {{\lbrace #1 \rbrace}}
\newcommand{\floor      } [1] {{\lfloor #1 \rfloor}}
\newcommand{\ceil       } [1] {{\lceil  #1 \rceil }}
\newcommand{\inverse    } [1] {{#1}^{-1}}
\newcommand{\transpose  } [1] {{#1}^\mathrm{T}}
\newcommand{\invtransp  } [1] {{#1}^{-\mathrm{T}}}
\newcommand{\relu       } [1] {{\lbrack #1 \rbrack_+}}

\newcommand{\abs        } [1] {{| #1 |}}
\newcommand{\Abs        } [1] {{\left| #1 \right|}}
\newcommand{\norm       } [1] {{\| #1 \|}}
\newcommand{\Norm       } [1] {{\left\| #1 \right\|}}
\newcommand{\pnorm      } [2] {\norm{#1}_{#2}}
\newcommand{\Pnorm      } [2] {\Norm{#1}_{#2}}
\newcommand{\inner      } [2] {{\langle {#1} \, | \, {#2} \rangle}}
\newcommand{\Inner      } [2] {{\left\langle \begin{array}{@{}c|c@{}}
                               \displaystyle {#1} & \displaystyle {#2}
                               \end{array} \right\rangle}}

\newcommand{\twopartdef}[4]
{
  \left\{
  \begin{array}{ll}
    #1 & \mbox{if } #2 \\
    #3 & \mbox{if } #4
  \end{array}
  \right.
}

\newcommand{\fourpartdef}[8]
{
  \left\{
  \begin{array}{ll}
    #1 & \mbox{if } #2 \\
    #3 & \mbox{if } #4 \\
    #5 & \mbox{if } #6 \\
    #7 & \mbox{if } #8
  \end{array}
  \right.
}

\newcommand{\len}[1]{\text{len}(#1)}

\newlength{\w}
\newlength{\h}
\newlength{\x}

\definecolor{darkred}{rgb}{0.7,0.1,0.1}
\definecolor{darkgreen}{rgb}{0.1,0.5,0.1}
\definecolor{cyan}{rgb}{0.7,0.0,0.7}
\definecolor{otherblue}{rgb}{0.1,0.4,0.8}
\definecolor{maroon}{rgb}{0.76,.13,.28}
\definecolor{burntorange}{rgb}{0.81,.33,0}

\ifdefined\ShowNotes
  \newcommand{\colornote}[3]{{\color{#1}\bf{#2 #3}\normalfont}}
\else
  \newcommand{\colornote}[3]{}
\fi

\newcommand {\note}[1]{\colornote{maroon}{}{#1}}
\newcommand {\todo}[1]{\colornote{cyan}{TODO}{#1}}
\newcommand {\ohad}[1]{\colornote{blue}{OF:}{#1}}
\newcommand {\maneesh}[1]{\colornote{red}{MA:}{#1}}
\newcommand {\MZ}[1]{\colornote{darkgreen}{MZ:}{#1}}
\newcommand {\eli}[1]{\colornote{otherblue}{ES:}{#1}}
\newcommand {\CT}[1]{\colornote{cyan}{CT:}{#1}}
\newcommand {\Adam}[1]{\colornote{darkgreen}{AF:}{#1}}
\newcommand {\AT}[1]{\colornote{burntorange}{AT:}{#1}}
\newcommand {\zeyu}[1]{\colornote{orange}{ZJ:}{#1}}
\newcommand {\dgo}[1]{\colornote{red}{A3:}{#1}}

\newcommand {\new}[1]{\colornote{red}{#1}}

\newcommand*\rot[1]{\rotatebox{90}{#1}}

\newcommand {\newstuff}[1]{#1}

\newcommand\todosilent[1]{}

\newcommand{\woBGmask}{{w/o~bg~\&~mask}}
\newcommand{\woMask}{{w/o~mask}}

\begin{abstract}
Editing talking-head video to change the speech content or to remove filler words is challenging.
We propose a novel method to edit talking-head video based on its transcript to produce a realistic output video in which the dialogue of the speaker has been modified, while maintaining a seamless audio-visual flow (i.e. no jump cuts).
Our method automatically annotates an input talking-head video 
with phonemes, visemes, 3D face pose and geometry, reflectance, expression and scene illumination per frame. 
To edit a video, the user has to only edit the transcript, and an optimization strategy then chooses segments of the input corpus as base material.
The annotated parameters corresponding to the selected segments are seamlessly stitched together and used to produce an intermediate video representation in which the lower half of the face is rendered with a parametric face model.
Finally, a recurrent video generation network transforms this representation to a photorealistic video that matches the edited transcript.
We demonstrate a large variety of edits, such as the addition, removal, and alteration of words, as well as convincing language translation and full sentence synthesis.
\end{abstract}


\setcopyright{acmlicensed}
\acmJournal{TOG}
\acmYear{2019}\acmVolume{38}\acmNumber{4}\acmArticle{68}\acmMonth{7} \acmDOI{10.1145/3306346.3323028}


\begin{teaserfigure}
\centering      
\includegraphics[width=\textwidth]{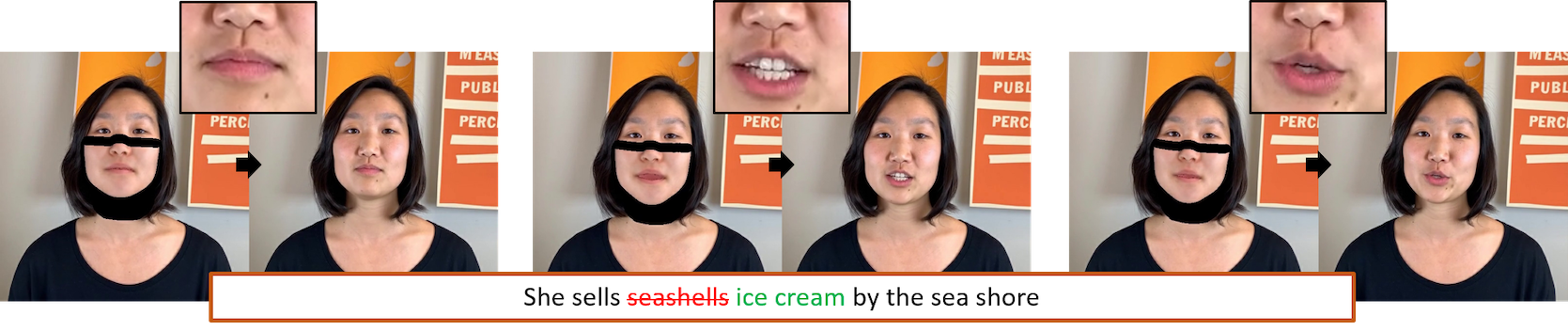}
\caption
{
	We propose a novel text-based editing approach for talking-head video.
	Given an edited transcript, our approach produces a realistic output video in which the dialogue of the speaker has been modified and the resulting video maintains a seamless audio-visual flow (i.e. no jump cuts).	
}
\label{fig:teaser}
\end{teaserfigure}

\maketitle

\section{Introduction}

Talking-head video -- framed to focus on the face and upper body of a speaker -- is ubiquitous in movies, TV shows, commercials, YouTube video logs, and online lectures.
Editing such pre-recorded video is challenging, 
but can be needed to emphasize particular content, remove filler words,
correct mistakes, or more generally match the editor's intent.
Using current video editing tools, like Adobe Premiere, skilled 
editors typically scrub through raw video footage to find relevant 
segments and assemble them into the desired story.
They must carefully consider where to place cuts so as to minimize
disruptions to the overall audio-visual flow.

Berthouzoz~\etal~\shortcite{Berthouzoz:2012:TPC:2185520.2185563} 
introduce a text-based approach for editing such videos.
Given an input video, they obtain a time-aligned transcript and allow
editors to cut and paste the text to assemble it into the
desired story.
Their approach can move or delete
segments, while generating visually seamless transitions 
at cut boundaries.
However, this method only produces artifact-free results when these 
boundaries are constrained to certain well-behaved segments of the video 
(e.g. where the person sits still between phrases or sentences).

Neither conventional editing tools nor the text-based approach
allow synthesis of new audio-visual speech content.
Thus, some modifications require either re-shooting the footage 
or overdubbing existing footage with new wording.
Both methods are expensive as they require new performances, and
overdubbing generally produces mismatches between the visible 
lip motion and the audio.

This paper presents a method that completes the suite of operations
necessary for transcript-based editing of talking-head video.
Specifically, based only 
on text edits, it can synthesize convincing new video of a person speaking, and produce seamless transitions even at challenging
cut points such as the middle of an utterance.

Our approach builds on a thread of research for synthesizing realistic 
talking-head video.
The seminal Video Rewrite system of 
Bregler~\etal~\shortcite{Bregler:1997:VRD:258734.258880} 
and the recent Synthesizing Obama project of 
Suwajanakorn~\etal~\shortcite{Suwajanakorn:2017:SOL:3072959.3073640} 
take new speech recordings as input, and 
superimpose the corresponding lip motion over talking-head video.
While the latter state-of-the art approach 
can synthesize fairly accurate lip sync, 
it has been shown to work for exactly one talking head because it requires \emph{huge} training data (14 hours). This method also relies on input audio from the same voice on which it was trained -- from either Obama or a voice impersonator. In contrast our approach works from text and therefore supports applications that require 
a different voice, such as translation.

Performance-driven puppeteering and dubbing methods, such as VDub\,\cite{GarriVSSVPT2015}, Face2Face\, \cite{ThiesZSTN2016a} and Deep Video Portraits\,\cite{kim2018DeepVideo}, take a new talking-head performance (usually from a different performer) as input and transfer the lip and head motion to the original talking-head video.
Because these methods have access to video as input they can often
produce higher-quality synthesis results than the audio-only methods.
Nevertheless, capturing new video for this purpose is obviously
more onerous than typing new text.

Our method accepts text only as input for synthesis, yet builds on the Deep Video Portraits approach of Kim~\etal~\shortcite{kim2018DeepVideo} to craft synthetic video.
Our approach drives a 3D model by seamlessly stitching different snippets of motion tracked
from the original footage. The snippets are selected based on a dynamic programming optimization
that searches for sequences of sounds in the transcript that should \emph{look} like the
words we want to synthesize, using a novel \emph{viseme}-based similarity measure.
These snippets can be re-timed to match the target viseme sequence,
and are blended to create a seamless mouth motion.
To synthesize output video, we first create a synthetic composite video in which the lower face region is masked out. In cases of inserting new text, we retime the rest of the face and background from the boundaries. The masked out region is composited with a synthetic 3D face model rendering using the mouth motion found earlier by optimization 
(\Cref{fig:data}). 
The composite exhibits the desired motion, but lacks realism due to the incompleteness and imperfections of the 3D face model. For example, facial appearance does not perfectly match, dynamic high-frequency detail is missing, and the mouth interior is absent.
Nonetheless, these data are sufficient cues for a new learned recurrent video generation network to be able to 
convert them to realistic imagery. The new composite representation and the recurrent network formulation significantly extend the neural face translation approach of Kim~\etal~\shortcite{kim2018DeepVideo} to text-based editing of existing videos. 

We show a variety of text-based editing results and favorable comparisons to previous techniques. In a crowd-sourced user study, our edits were rated to be real in 59.6\% of cases.
The main technical contributions of our approach are:
\begin{itemize}
	\item A text-based editing tool for talking-head video that lets editors insert new text, in addition to cutting and copy-pasting in an existing transcript.
	\item A dynamic programming based strategy tailored to video synthesis that assembles new words based on snippets containing sequences of observed visemes in the input video.
	\item A parameter blending scheme that, when combined with our synthesis pipeline, produces seamless talking heads, even when combining snippets with different pose and expression.
	\item A recurrent video generation network that converts a composite of real background video and synthetically rendered lower face into a photorealistic video. 
\end{itemize}

\subsection{Ethical Considerations}
\label{sec:social}

\newstuff{
Our text-based editing approach lays the foundation for better editing
tools for movie post production. Filmed dialogue scenes often require
re-timing or editing based on small script changes, which currently
requires tedious manual work. Our editing technique also enables easy
adaptation of audio-visual video content to specific target audiences:
e.g., instruction videos can be fine-tuned to audiences of different
backgrounds, or a storyteller video can be adapted to children of
different age groups purely based on textual script edits. In short,
our work was developed for storytelling purposes.

However, the availability of such technology --- at a quality that some
might find indistinguishable from source material --- also raises
important and valid concerns about the potential for misuse. Although
methods for image and video manipulation are as old as the media
themselves, the risks of abuse are heightened when applied to a mode
of communication that is sometimes considered to be authoritative
evidence of thoughts and intents. We acknowledge that bad actors might
use such technologies to falsify personal statements and slander
prominent individuals. We are concerned about such deception and misuse.

Therefore, we believe it is critical that video synthesized using our tool
clearly presents itself as synthetic. 
The fact that the video is
synthesized may be obvious by context (e.g. if the audience
understands they are watching a fictional movie), directly stated in
the video or signaled via watermarking. We also believe that it is
essential to obtain permission from the performers for any alteration
before sharing a resulting video with a broad audience. Finally, it is
important that we as a community continue to develop forensics,
fingerprinting and verification techniques (digital and non-digital)
to identify manipulated video. Such safeguarding measures would reduce
the potential for misuse while allowing creative uses of video editing
technologies like ours.

We hope that publication of the technical details of such systems can
spread awareness and knowledge regarding their inner workings,
sparking and enabling associated research into the aforementioned
forgery detection, watermarking and verification systems. Finally, we
believe that a robust public conversation is necessary to create a set
of appropriate regulations and laws that would balance the risks of
misuse of these tools against the importance of creative, consensual
use cases.
}

\begin{figure*}[ht]
	\includegraphics[width=0.95\textwidth]{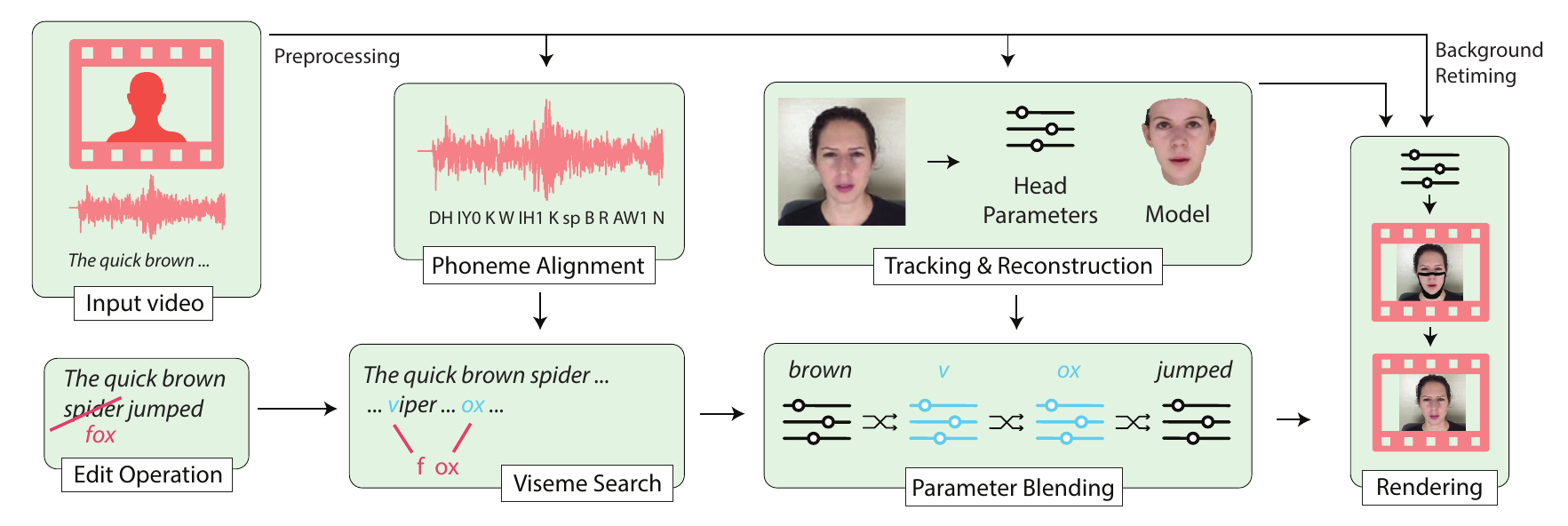}
	\vspace{-0.3cm}
	\caption{Method overview. Given an input talking-head video and a transcript, we perform text-based editing. We first align phonemes to the input audio and track each input frame to construct a parametric head model. Then, for a given edit operation (changing \emph{spider} to \emph{fox}), we find segments of the input video that have similar visemes to the new word. In the above case we use \emph{\textbf{v}iper} and \emph{\textbf{ox}} to construct \emph{fox}. We use blended head parameters from the corresponding video frames, together with a retimed background sequence, to generate a composite image, which is used to generate a photorealistic frame using our neural face rendering method. In the resulting video, the actress appears to be saying \emph{fox}, even though that word was never spoken by her in the original recording.
	}
	\label{fig:overview}
\end{figure*}

\section{Related Work} \label{sec:related}

\paragraph{Facial Reenactment.}
Facial video reenactment has been an active area of  research\,\cite{LiuSZ2001,Suwajanakorn:2017:SOL:3072959.3073640,AverbCKC2017,VlasiBPP2005,KemelSSS2010,GarriVRTPT2014,LiDWLXW2014}.
Thies et al.\,\shortcite{ThiesZSTN2016a} recently demonstrated real-time video reenactment.
Deep video portraits\,\cite{kim2018DeepVideo} enables full control of the head pose, expression, and eye gaze of a target actor based on recent advances in learning-based image-to-image translation~ \cite{IsolaZZE2017}.
Some recent approaches enable the synthesis of controllable facial animations from single images \cite{AverbCKC2017,Geng:2018,Wiles18}.
Nagano et al.\,\shortcite{Nagano:2018} recently showed 
how to estimate a controllable avatar of a person from a single image. 
We employ a facial reenactment approach for visualizing our text-based editing results and show how facial reenactment can be tackled by neural face rendering.

\paragraph{Visual Dubbing.}
Facial reenactment is the basis for visual dubbing, since it allows to alter the expression of a target actor to match the motion of a dubbing actor that speaks in a different language.
Some dubbing approaches are speech-driven\,\cite{Bregler:1997:VRD:258734.258880,ChangE2005,EzzatGP2002,LiuO2011} others are performance-driven\,\cite{GarriVSSVPT2015}.
Speech-driven approaches have been shown to produce accurate lip-synced video\,\cite{Suwajanakorn:2017:SOL:3072959.3073640}.
While this approach can synthesize fairly accurate lip-synced video, it requires the new audio to sound similar to the original speaker, while we enable synthesis of new video using text-based edits. 
\newstuff{
Mattheyses et al.\,\shortcite{DBLP:conf/avsp/MattheysesLV10} show results with no head motion, in a controlled setup with uniform background. In contrast, our 3D based approach and neural renderer can produce subtle phenomena such as lip rolling, and works in a more general setting.
}

\newstuff{
\paragraph{Speech animation for rigged models.}
Several related methods
produce animation curves for speech %
\cite{Edwards:2016:JAV:2897824.2925984,Taylor:2017:DLA:3072959.3073699,Zhou:2018:VAA:3197517.3201292}.
They are specifically designed for animated 3D models and not for photorealistic video,
requiring a character rig and artist supplied rig correspondence. 
In contrast, our approach ``animates'' a real person speaking, based just on text and a monocular recording of the subject.
}

\paragraph{Text-Based Video and Audio Editing.}
Researchers have developed a variety of audio and video editing tools based on time-aligned transcripts.
These tools allow editors to shorten and rearrange speech for audio podcasts\,\cite{rubin2013,shin2016}, annotate video with review feedback\,\cite{pavel2016}, provide audio descriptions of the video content for segmentation of B-roll footage\,\cite{truong2016} and generate structured summaries of lecture videos\,\cite{pavel2014}.
Leake et al.\,\shortcite{Leake:2017:CVE:3072959.3073653} use the structure imposed by time-aligned transcripts to automatically edit together multiple takes of a scripted scene based on higher-level cinematic idioms specified by the editor.
Berthouzoz et al.'s\,\shortcite{Berthouzoz:2012:TPC:2185520.2185563} tool for editing interview-style talking-head video by cutting, copying and pasting transcript text is closest to our work.
While we similarly enable rearranging video by cutting, copying and pasting text, unlike all of the previous text-based editing tools, we allow synthesis of new video by simply typing the new text into the transcript.

\paragraph{Audio Synthesis.}

In transcript-based video editing, synthesizing new video clips would
often naturally be accompanied by audio synthesis. Our approach to
video is independent of the audio, and therefore a variety of
\emph{text to speech} (TTS) methods can be used.
Traditional TTS has explored two general approaches:
\emph{parametric methods} (\eg~\cite{zen2009statistical}) generate
acoustic features based on text, and then synthesize a waveform from
these features. Due to oversimplified acoustic models, they tend to
sound robotic. In contrast, 
\emph{unit selection} is a data driven approach that constructs new waveforms by stitching together small pieces of audio (or \emph{units}) found elsewhere in the transcript~\cite{hunt1996unit}. 
Inspired by the latter, the VoCo project of
Jin~\etal~\shortcite{jin2017voco} performs a search in the existing
recording to find short ranges of audio that can be stitched together
such that they blend seamlessly in the context around an insertion
point. Section~\ref{sec:applications} and the accompanying video
present a few examples of using our method to synthesize new words in video, 
coupled with the use of VoCo to synthesize corresponding audio.
Current state-of-the-art TTS approaches rely on deep learning\,\cite{van2016wavenet,shen2018natural}.
However, these methods require a huge (tens of hours) training corpus for the target
speaker.

\paragraph{Deep Generative Models.}
Very recently, researchers have proposed \textsl{Deep Generative Adversarial Networks (GANs)} for the synthesis of images and videos.
Approaches create new images from scratch \,\cite{GoodfPMXWOCB2014,RadfoMC2016,ChenK2017,KarraALL2018,WangLZTKC2018} or condition the synthesis on an input image \,\cite{MirzaO2014,IsolaZZE2017}.
High-resolution conditional video synthesis\,\cite{wang2018vid2vid} has recently been demonstrated.
Besides approaches that require a paired training corpus, unpaired video-to-video translation techniques\,\cite{Recycle-GAN} only require two training videos.
Video-to-video translation has been used in many applications.
For example, impressive results have been shown for the reenactment of the human head \cite{OlszeLYZYHXSKL2017},
head and upper body \cite{kim2018DeepVideo}, and the whole human body\,\cite{Chan2018,Liu2018}.

\paragraph{Monocular 3D Face Reconstruction.}
There is a large body of work on reconstructing facial geometry and appearance from a single image using optimization
methods\,\cite{Kemel2013,RothTL2017,ShiWTC2014,GarriZCVVPT2016,ThiesZSTN2016a,Suwajanakorn:2017:SOL:3072959.3073640,FyffeJAID2014,IchimBP2015}.
Many of these techniques employ a parametric face model\,\cite{BlanzV1999,BlanzSVS2004,BoothRPDZ2018} as a prior to better constrain the reconstruction problem.
Recently, deep learning-based approaches have been proposed that train a convolutional network to directly regress the model parameters\,\cite{TewarZKGBPT2017,8496850,TranHMM2017,RichaSK2016,Genova_2018_CVPR, Dou_2017_CVPR}.
Besides model parameters, other approaches regress detailed depth maps\,\cite{SelaRK2017,RichaSOK2017}, or 3D displacements\,\cite{CaoBZB2015,8360505, tewari2017self}.
Face reconstruction is the basis for a large variety of applications, such as facial reenactment and visual dubbing.
For more details on monocular 3D face reconstruction, we refer to Zollh\"{o}fer et al.\,\shortcite{Zollhoefer2018FaceSTAR}.

\section{Method}
\label{sec:method}

Our system takes as input a video recording of a talking head with a
transcript of the speech and any number of edit operations specified
on the transcript.  Our tool supports three types of edit operations;
\begin{itemize}
\item{\bf \em Add new words:} the edit adds one or more consecutive words at a point in the video (e.g. because the actor
skipped a word or the producer wants to insert a phrase).
 \item{\bf \em Rearrange existing words:} the edit moves one or more consecutive words that exist in the video (e.g. for better word
ordering without introducing jump cuts).
 \item{\bf \em Delete existing words:} the edit removes one or more consecutive words from the video (e.g. for simplification of wording and removing filler such as ``um'' or ``uh'').
\end{itemize}
We represent editing operations by the sequence of words $\EditOp$
in the edited region as well as the correspondence between those words
and the original transcript.
For example, deleting the word ``wonderful'' in the sequence ``hello
wonderful world'' is specified as (`hello', `world') and adding the
word ``big'' is specified as (`hello', `big', `world').

Our system processes these inputs in five main stages
(\Cref{fig:overview}). 
In the phoneme alignment stage (\Cref{subsec:phoneme_align}) we align
the transcript to the video at the level of phonemes and then in the
tracking and reconstruction stage (\Cref{subsec:face_reconstruction})
we register a 3D parametric head model with the video. These are
pre-processing steps performed once per input video.
Then for each edit operation $\EditOp$ we first perform a viseme search
(\Cref{subsec:viseme_search}) to find the best visual match between
the subsequences of phonemes in the edit and subsequences of phonemes
in the input video. 
We also extract a region around the edit location to act as a background sequence, from which we will extract background pixels and pose data.
For each subsequence we blend the parameters
of the tracked 3D head model (\Cref{subsec:param_blend}) and then use
the resulting parameter blended animation of the 3D head, together with the background pixels, to render a
realistic full-frame video (\Cref{subsec:rendering}) in which the
subject appears to say the edited sequence of words. 
Our viseme search and approach for combining shorter subsequences with parameter blending is motivated by the phoneme/viseme distribution of the English language (\Cref{subsec:data_analysis}).

\subsection{Phoneme Alignment}
\label{subsec:phoneme_align}

Phonemes are perceptually distinct units that distinguish one
word from another \newstuff{in a specific language}.
Our method relies on phonemes to find snippets in the video that we
later combine to produce new content. Thus, our first step is to
compute the \emph{identity} and \emph{timing} of phonemes in the input
video.
To segment the video's speech audio into \newstuff{phones (audible realizations of phonemes)}, we assume we have
an accurate text transcript 
and align it to the
audio using P2FA~\cite{yuan2008,rubin2013}, a phoneme-based alignment
tool.  This gives us an ordered sequence $\PhoneSeq = (\Phone_1, \ldots, \Phone_n)$ of
phonemes, each with a label denoting the phoneme name, start time, and
end time $\Phone_i = (\Phone_i^\textit{lbl}, \Phone_i^\textit{in},
\Phone_i^\textit{out})$. Note that if a transcript is not given as part of the
input, we can use automatic speech transcription
tools~\cite{ibmwatson,gentle} or crowdsourcing transcription services
like \texttt{\url{rev.com}} to obtain it.

\subsection{3D Face Tracking and Reconstruction}
\label{subsec:face_reconstruction}
We register a 3D parametric face model with each frame of the input
talking-head video.  The parameters of the model (e.g.
expression, head pose, etc.)  will later allow us to selectively blend
different aspects of the face (e.g. take the expression from one frame and
pose from another).  Specifically, we apply recent work on monocular
model-based face
reconstruction\,\cite{GarriZCVVPT2016,ThiesZSTN2016a}. These techniques parameterize
the rigid head pose $\mathbf{T} \in SE(3)$, the facial geometry $\boldsymbol\alpha \in \mathbb{R}^{80}$, facial reflectance $\boldsymbol\beta \in \mathbb{R}^{80}$, facial expression $\boldsymbol\delta \in \mathbb{R}^{64}$, and scene illumination $\boldsymbol\gamma \in \mathbb{R}^{27}$.
Model fitting is based on the minimization of a non-linear
reconstruction energy.
For more details on the minimization, please see the papers of \citet{GarriZCVVPT2016} and \citet{ThiesZSTN2016a}.
In total, we obtain a $257$ parameter vector $\mathbf{p} \in
\mathbb{R}^{257}$
for each frame of the input video.

\begin{figure*}
    \setlength\tabcolsep{1.5pt}
    \newlength{\mycolw}
    \setlength{\mycolw}{0.1\textwidth}
    	\vspace{-0.1in}
    \begin{tabular}{cc|c|cc|cc|ccc}
        &
        OF &
        FRE &
        \multicolumn{2}{|c|}{NCH T} &
        \multicolumn{2}{|c|}{OAST} &
        IN & & \\
    
       \raisebox{0.9\normalbaselineskip}[0pt][0pt]{\rot{Original}} &
    	\includegraphics[width=\mycolw]{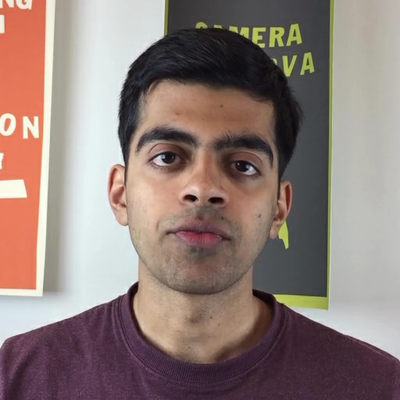} &
    	\includegraphics[width=\mycolw,cfbox=blue 1pt 1pt]{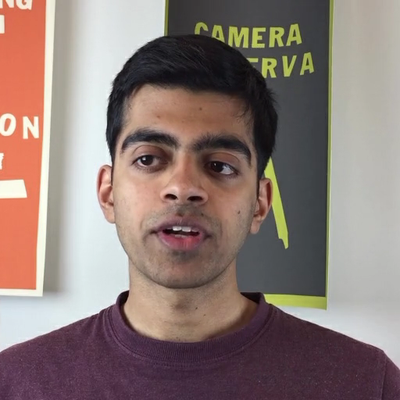} &
    	\includegraphics[width=\mycolw,cfbox=blue 1pt 1pt]{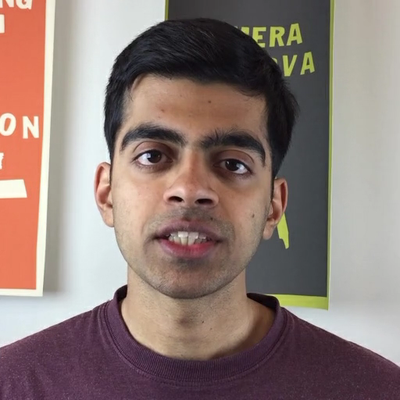} &
    	\includegraphics[width=\mycolw]{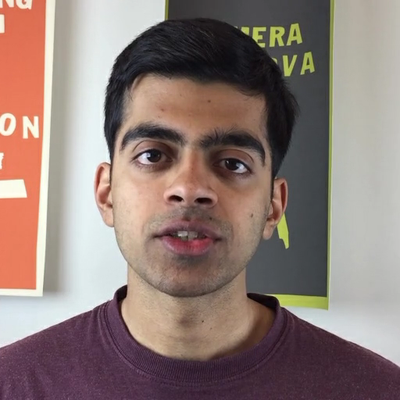} &
    	\includegraphics[width=\mycolw]{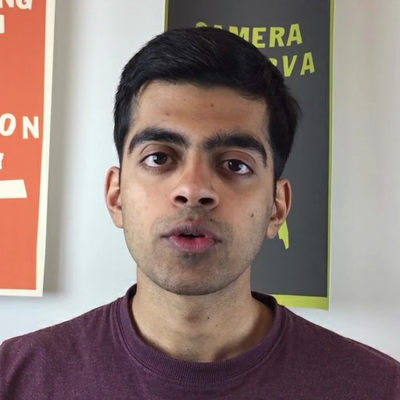} &
    	\includegraphics[width=\mycolw]{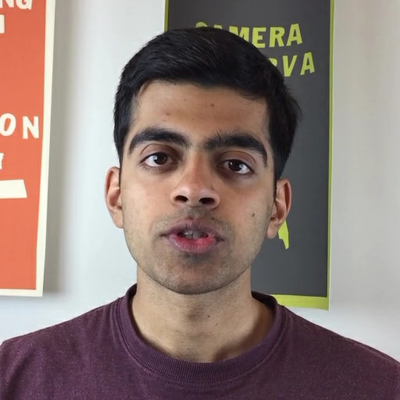} &
    	\includegraphics[width=\mycolw]{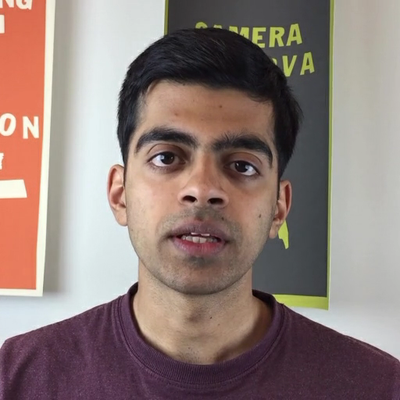} & &
    	\includegraphics[width=\mycolw,cfbox=blue 1pt 1pt]{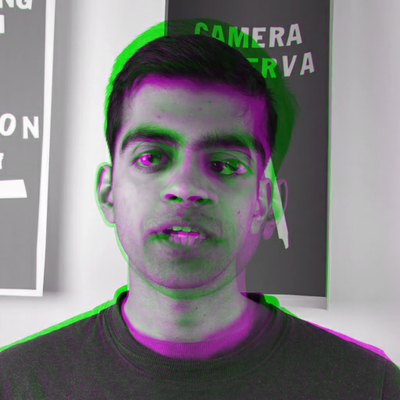} \\
    	
    	\raisebox{1.2\normalbaselineskip}[0pt][0pt]{\rot{Render}} &
    	\includegraphics[width=\mycolw]{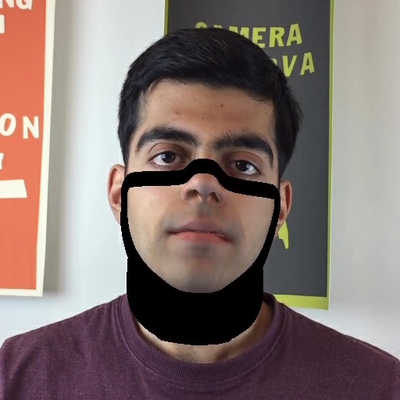} &
    	\includegraphics[width=\mycolw]{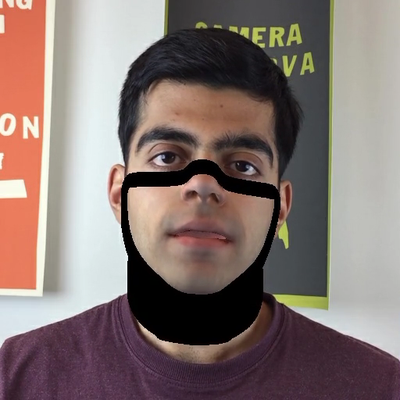} &
    	\includegraphics[width=\mycolw]{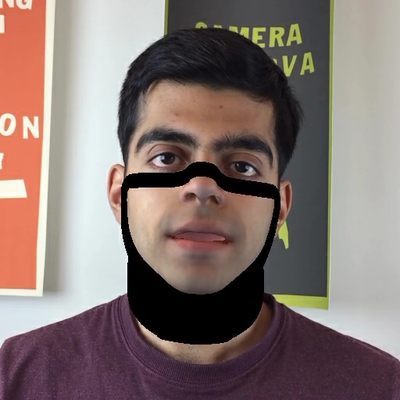} &
    	\includegraphics[width=\mycolw]{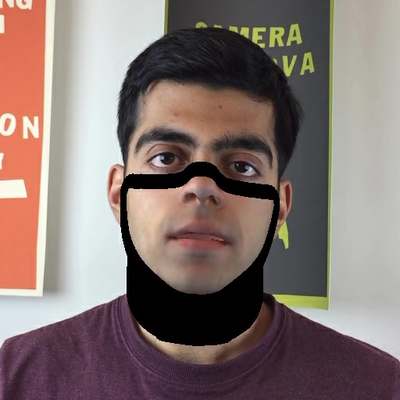} &
    	\includegraphics[width=\mycolw]{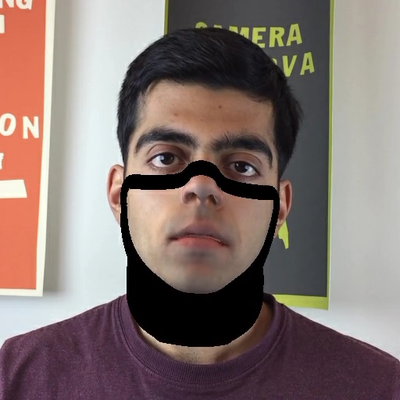} &
    	\includegraphics[width=\mycolw]{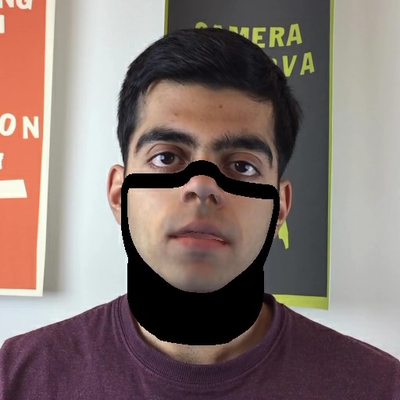} &
    	\includegraphics[width=\mycolw]{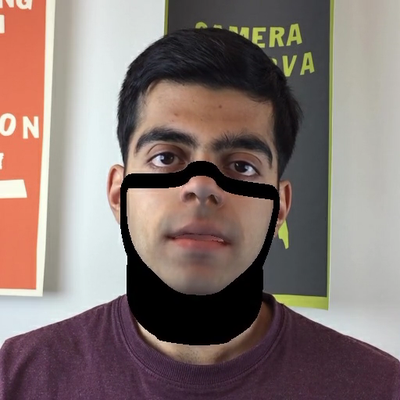} & & \\
    	
    	\raisebox{.3\normalbaselineskip}[0pt][0pt]{\rot{Synthesized}} &
    	\includegraphics[width=\mycolw]{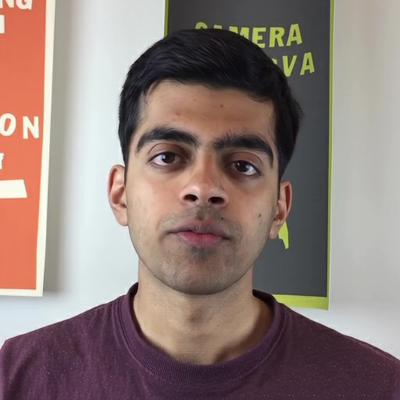} &
    	\includegraphics[width=\mycolw,cfbox=red 1pt 1pt]{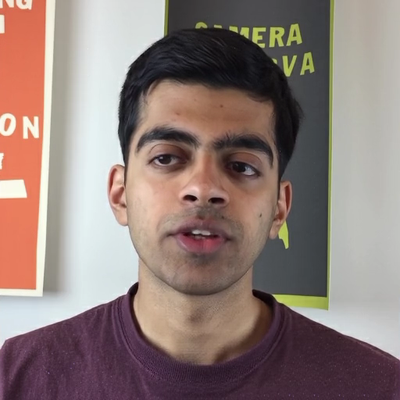} &
    	\includegraphics[width=\mycolw,cfbox=red 1pt 1pt]{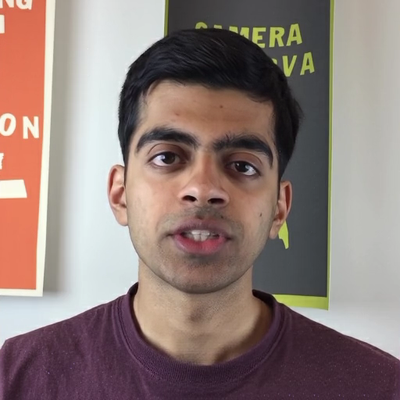} &
    	\includegraphics[width=\mycolw]{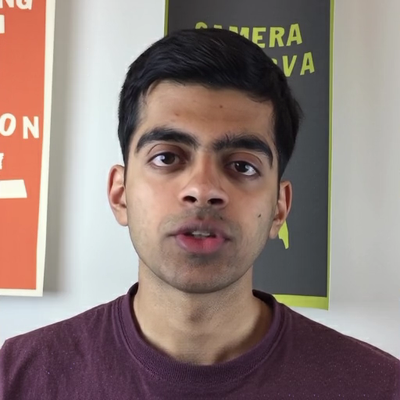} &
    	\includegraphics[width=\mycolw]{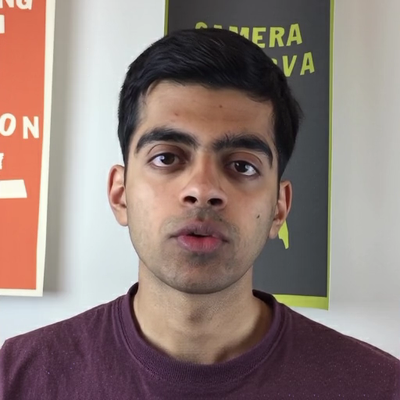} &
    	\includegraphics[width=\mycolw]{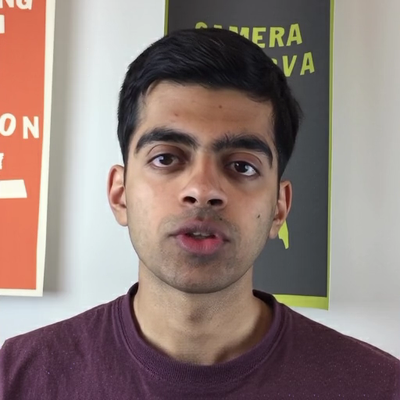} &
    	\includegraphics[width=\mycolw]{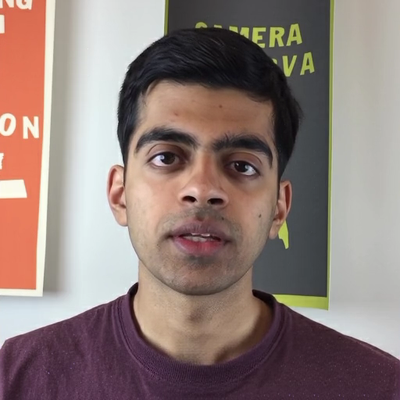} & & 
    	\includegraphics[width=\mycolw,cfbox=red 1pt 1pt]{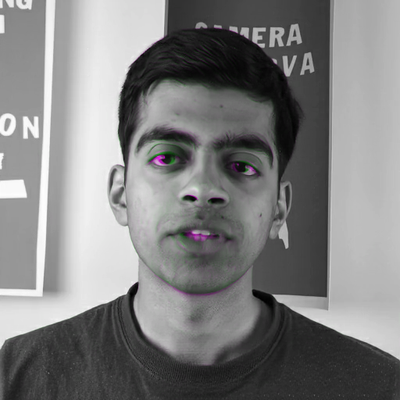} \\
    \end{tabular}

	\vspace{-0.1in}
	\caption
	{
	    Our parameter blending strategy produces a seamless synthesized result from choppy original sequences. Above, we insert the expression ``french toast'' instead of ``napalm'' in the sentence ``I like the smell of napalm in the morning.'' The new sequence was taken from different parts of the original video:
		F R EH1 taken from ``fresh'', 
		N CH T taken from ``drenched'', and
		OW1 S T taken from ``roast''.
		Notice how original frames from different sub-sequences are different in head size and posture, while our synthesized result is a smooth sequence. On the right we show the pixel difference between blue and red frames; notice how blue frames are very different. \textit{Videos in supplemental material. }
               	\vspace{-0.05in}
	}
	\label{fig:blending1}
\end{figure*}

\subsection{Viseme Search}
\label{subsec:viseme_search}

\begin{figure}
	\includegraphics[width=0.9\columnwidth]{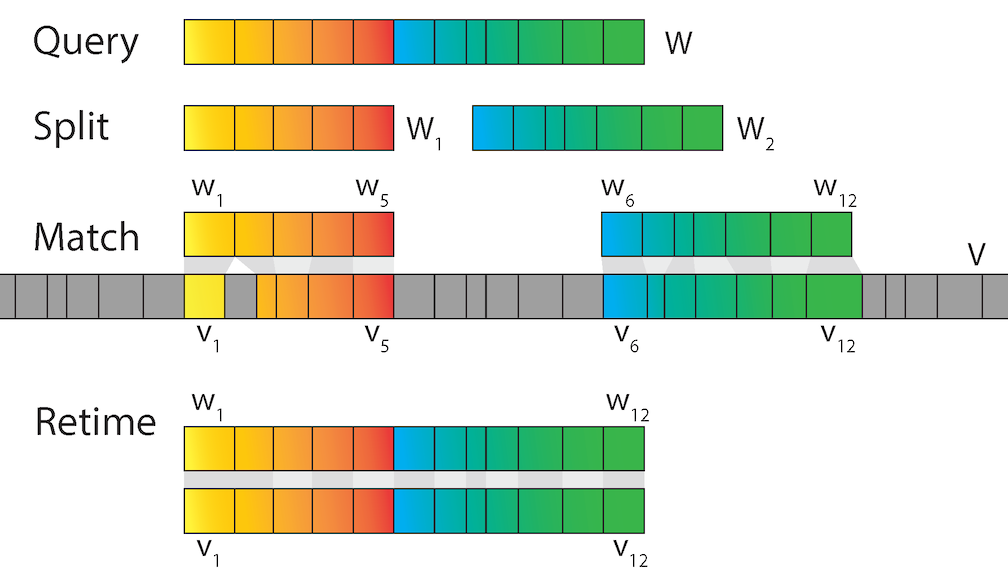}\\
	\vspace{-0.3cm}
	\caption
	{
		Viseme search and retiming. Given a query sequence $\QuerySeq$,
		We split it into all possible subsequences, of which one $(\QuerySeq_1, \QuerySeq_2) \in \Split(\QuerySeq)$ is shown.
		Each subsequence is matched to the input video $\PhoneSeq$, producing a correspondance between query phonemes $\Query_{i}$ and input video phonemes $\Phone_{i}$. We retime in parameter space to match the lengths of each $\Phone_i$ to $\Query_i$.
	}
	\label{fig:search}
	\vspace{-0.4cm}
\end{figure}

Given an edit operation specified as a sequence of words $\EditOp$,
our goal is to find matching sequences of phonemes in the video
that can be combined to produce $\EditOp$. 
In the
matching procedure we use the fact that 
identical phonemes are expected to be, on average, more visually similar to each other than non-identical phonemes (despite co-articulation effects).
We similarly consider visemes, groups of aurally distinct
phonemes that appear visually similar to one another
(\Cref{tbl:visemes}), as good potential matches.  Importantly, the
matching procedure \emph{cannot} expect to find a good coherent viseme
sequence in the video for long words or sequences in the edit
operation. Instead, we must find several matching subsequences and a
way to best combine them.

We first convert the edit operation $\EditOp$ to a phoneme sequence
$\QuerySeq = (\Query_1, \ldots, \Query_m)$
where each $\Query_i$ is defined as $(\Query_i^\textit{lbl},
\Query_i^\textit{in}, \Query_i^\textit{out})$ similar to our definition of phonemes in the video $\Phone_i$.  
We can convert the text $\EditOp$ to phoneme labels $\Query_i^\textit{lbl}$ using a word to phoneme map, but text does not contain timing information $\Query_i^\textit{in}, \Query_i^\textit{out}$.
To obtain timings we use a text-to-speech synthesizer to convert the edit into speech. For all results in this paper we use either the built-in speech synthesizer in Mac OS X, or Voco\,\cite{jin2017voco}. Note however that our video synthesis pipeline does not use the audio \emph{signal}, but only its \emph{timing}. So, e.g., manually specified \newstuff{phone} lengths could be used as an alternative.
The video generated in the rendering stage of our pipeline (\Cref{subsec:rendering}) is mute and we discuss how we can add audio at the end of that section.
Given the audio of $\EditOp$, we produce phoneme labels and timing using P2FA, in a
manner similar to the one we used in \Cref{subsec:phoneme_align}.

Given an edit $\QuerySeq$ 
and the video phonemes $\PhoneSeq$, we are looking for the optimal partition of $\QuerySeq$ into sequential subsequences $\QuerySeq_1, \dots, \QuerySeq_k$, such that each subsequence has a good match in $\PhoneSeq$, while encouraging subsequences to be long (\Cref{fig:search}). 
We are looking for long subsequences because each transition between subsequences may cause artifacts in later stages.
We first describe matching one subsequence $\QuerySeq_i = (\Query_{j}, \dots, \Query_{j+k})$ to the recording $\PhoneSeq$, and then explain how we match the full query $\QuerySeq$.

\paragraph{Matching one subsequence}
We define $\CostMatch(\QuerySeq_i, \SomePhonemeSubseq)$ between a
subsequence of the query $\QuerySeq_i$ and some subsequence of the
video $\SomePhonemeSubseq$ as a modified Levenshtein edit distance
\cite{levenshtein1966binary} between phoneme
sequences that takes phoneme length into account.
The edit distance requires pre-defined costs for insertion, deletion and swap.
We define our insertion cost $\CostInsert = 1$ and deletion cost
$\CostDelete = 1$ and consider viseme and phoneme labels as well as phoneme lengths in our swap cost
\begin{equation}
    \label{eqn:c_swap}
    \CostSwap(\Phone_i, \Query_j) = \CostViseme(\Phone_i, \Query_j) (\abs{\Phone_i} + \abs{\Query_j}) + \chi\Abs{\abs{\Phone_i} - \abs{\Query_j}}
\end{equation}
where $\abs{a}$ denotes the length of phoneme $a$, 
$\CostViseme(\Phone_i, \Query_j)$ is $0$ if $\Phone_i$ and $\Query_j$
are the same phoneme, $0.5$ if they are different phonemes but the
same viseme (\Cref{tbl:visemes}), and $1$ if they are different visemes.
The parameter $\chi$ controls the influence of length difference on the cost, and we set it to $10^{-4}$ in all our examples.
\Cref{eqn:c_swap} penalized for different phonemes and visemes,
weighted by the sum of the phoneme length. Thus longer non-matching
phonemes will incur a larger penalty, as they are more likely to be
noticed. 

We minimize $\CostMatch(\QuerySeq_i, \PhoneSeq)$ over all possible
$\SomePhonemeSubseq$ using dynamic programming
\cite{levenshtein1966binary} to find the best suffix of any prefix of
$\PhoneSeq$ and its matching cost to $\QuerySeq_i$. We brute-force all
possible prefixes of $\PhoneSeq$ to find the best match $\PhoneSeq_i$
to the query $\QuerySeq_i$.

\begin{table}
\label{tbl:visemes}
\begin{center}
\caption{Grouping phonemes (listed as ARPABET codes) into visemes.
 \newstuff{We use the viseme grouping of Annosoft's lipsync tool \cite{annosoft}. More viseme groups may lead to better visual matches (each group is more specific in its appearance), but require more data because the chance to find a viseme match decreases.
 We did not perform an extensive evaluation of different viseme groupings, of which there are many.}}
\vspace{-0.15cm}
\begin{tabular}{cp{0.48\linewidth}cp{0.26\linewidth}}
  \toprule
  \footnotesize{v01} & AA0, AA1, AA2                & \footnotesize{v09} & Y, IY0, IY1, IY2 \\
  \footnotesize{v02} & AH0, AH1, AH2, HH            & \footnotesize{v10} & R, ER0, ER1, ER2 \\
  \footnotesize{v03} & AO0, AO1, AO2                & \footnotesize{v11} & L \\
  \footnotesize{v04} & AW0, AW1, AW2, OW0,          & \footnotesize{v12} & W \\
                     & OW1, OW2                     & \footnotesize{v13} & M, P, B \\
  \footnotesize{v05} & OY0, OY1, OY2, UH0, UH1,     & \footnotesize{v14} & N, NG, DH, D, G, \\
                     & UH2, UW0, UW1, UW2           &                    & T, Z, ZH, TH, K, S \\
  \footnotesize{v06} & EH0, EH1, EH2, AE0, AE1, AE2 & \footnotesize{v15} & CH, JH, SH \\       
  \footnotesize{v07} & IH0, IH1, IH2, AY0, AY1, AY2 & \footnotesize{v16} & F, V \\
  \footnotesize{v08} & EY0, EY1, EY2                & \footnotesize{v17} & sp \\
  \bottomrule
\end{tabular}
\end{center}
\end{table}

\paragraph{Matching the full query}
We define our full matching cost $C$ between the query $\QuerySeq$ and the video $\PhoneSeq$ as
\begin{equation}
    \label{eqn:match_cost}
    C(\QuerySeq, \PhoneSeq) = \min_{\substack{(\QuerySeq_1, \dots, \QuerySeq_k) \in \Split(\QuerySeq) \\ (\PhoneSeq_1, \dots, \PhoneSeq_k) 
    }}{\sum_{i=1}^{k}{\CostMatch(\QuerySeq_i, \PhoneSeq_i) + \CostLength(\QuerySeq_i)}}
\end{equation}
where $\Split(\QuerySeq)$ denotes the set of all possible ways of splitting $\QuerySeq$ into subsequences,
and $\PhoneSeq_i$ is the best match for $\QuerySeq_i$ according to $\CostMatch$.
The cost $\CostLength(\QuerySeq_i)$ penalizes short subsequences and is defined as 
\begin{equation}
    \CostLength(\QuerySeq_i) = \frac{\phi}{\Abs{\QuerySeq_i}}
\end{equation}
where $\Abs{\QuerySeq_i}$ denotes the number of phonemes in subsequence $\QuerySeq_i$ and $\phi$ is a weight parameter empirically set to $0.001$ for all our examples. 
To minimize \Cref{eqn:match_cost} 
we generate all splits $(\QuerySeq_1, \dots, \QuerySeq_k) \in \Split(\QuerySeq)$ of the query (which is typically short), and for each $\QuerySeq_i$ we find the best subsequence $\PhoneSeq_i$ of $\PhoneSeq$ with respect to $\CostMatch$.
Since the same subsequence $\QuerySeq_i$ can appear in multiple partitions, we memoize computations to make sure each match cost is computed only once.
The viseme search procedure produces subsequences $(\PhoneSeq_1,
\dots, \PhoneSeq_k)$ of the input video that, when combined, should
produce $\QuerySeq$.

\subsection{Parameter Retiming \& Blending}
\label{subsec:param_blend}

The sequence $(\PhoneSeq_1, \dots, \PhoneSeq_k)$ of video subsequences describes sections of the video for
us to combine in order to create $\EditOp$. 
However, we cannot directly use the video frames that correspond to $(\PhoneSeq_1, \dots, \PhoneSeq_k)$ 
for two reasons:
(1) A sequence $\PhoneSeq_i$ corresponds to part of $\EditOp$ in viseme identity, but not in viseme length, which will produce unnatural videos when combined with the speech audio, 
and (2) Consecutive sequences $\PhoneSeq_i$ and $\PhoneSeq_{i+1}$ can be from sections that are far apart in the original video. The subject might look different in these parts due to pose and posture changes, movement of hair, or camera motion. Taken as-is, the transition between consecutive sequences will look unnatural (\Cref{fig:blending1} top).

To solve these issues, we use our parametric face model in order to
mix different properties (pose, expression, etc.) from
different input frames, and blend them in parameter space.
We also select a background sequence $\BackgroundSeq$ and
use it for pose data and background pixels. The background sequence allows us to edit challenging videos with hair movement and slight camera motion.

\paragraph{Background retiming and pose extraction}

An edit operation $\EditOp$ will often change the length of the
original video. We take a video sequence (from the input video)
$\BackgroundSeq'$ around the location of the edit operation, and
retime it to account for the change in length the operation will
produce, resulting in a retimed background sequence
$\BackgroundSeq$. We use nearest-neighbor sampling of frames, and
select a large enough region around the edit operation so that
retiming artifacts are negligible.
All edits in this paper use the length of one sentence as background.
The retimed sequence
$\BackgroundSeq$ does not match the original nor the new audio, but
can provide realistic background pixels and pose parameters that
seamlessly blend into the rest of the video. In a later step we synthesize frames based on the retimed background and expression parameters that \emph{do} match the audio.

\paragraph{Subsequence retiming}
The phonemes in each sequence $\Phone_{j} \in \PhoneSeq_i$ 
approximately match the length of corresponding query phonemes, but an exact match is required so that the audio and video will be properly synchronized. We set a desired frame rate $\mathcal{F}$ for our synthesized video, which often matches the input frame-rate, but does not have to (e.g. to produce slow-mo video from standard video). Given the frame rate $\mathcal{F}$, we sample 
model parameters $\mathbf{p}
\in \mathbb{R}^{257}$
by linearly interpolating adjacent frame parameters described in
\Cref{subsec:face_reconstruction}.
For each $\Phone_{j} \in \PhoneSeq_i$ we sample $\mathcal{F}\abs{\Query_{j}}$ %
frame parameters in $[\Phone_{j}^{in}, \Phone_{j}^{out}]$ so that the length of the generated video matches the length of the query $\abs{\Query_{j}}$. This produces a sequence 
that matches $\EditOp$ in timing, but with visible jump cuts between sequences if rendered as-is (\Cref{fig:search} bottom).

\paragraph{Parameter blending.}
To avoid jump cuts, we use different strategies for different parameters, as follows.
Identity geometry $\boldsymbol\alpha \in \mathbb{R}^{80}$ and reflectance $\boldsymbol\beta \in \mathbb{R}^{80}$ are kept constant throughout the sequence (it's always the same person), so they do not require blending.
Scene illumination $\boldsymbol\gamma \in \mathbb{R}^{27}$ typically changes slowly or is kept constant, thus we linearly interpolate illumination parameters between the last frame prior to the inserted sequence and the first frame after the sequence, disregarding the original illumination parameters of $\PhoneSeq_i$. This produces a realistic result while avoiding light flickering for input videos with changing lights.
Rigid head pose $\mathbf{T} \in SE(3)$ is taken directly from the
retimed background sequence $\BackgroundSeq$. This ensures that the
pose of the parameterized head model matches the background pixels in
each frame.

Facial expressions $\boldsymbol\delta \in \mathbb{R}^{64}$ are the
most important parameters for our task, as they hold information about
mouth and face movement --- the visemes we aim to reproduce. Our goal is
to preserve the retrieved expression parameters as much as possible,
while smoothing out the transition between
them. 
Our approach is to smooth out each
transition from $\PhoneSeq_i$ to $\PhoneSeq_{i+1}$ by linearly interpolating a region of
67 milliseconds around the transition. We found this length to be short enough so that individual \newstuff{visemes} %
are not lost, and long enough to produce convincing transitions
between \newstuff{visemes}.

\begin{figure}
  \centering
	\begin{subfigure}[b]{0.32\columnwidth}
		\includegraphics[width=\textwidth]{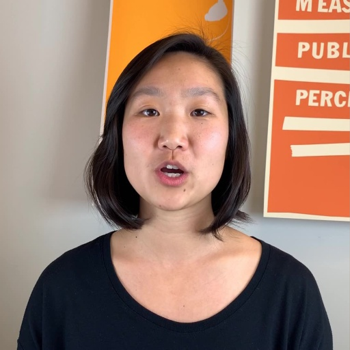}
		\caption{Ground Truth ($\mathbf{f}_i$)}
	\end{subfigure}
	\begin{subfigure}[b]{0.32\columnwidth}
		\includegraphics[width=0.49\textwidth]{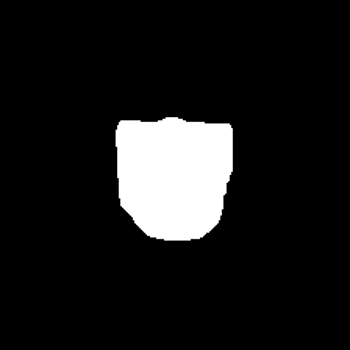}
		\includegraphics[width=0.49\textwidth]{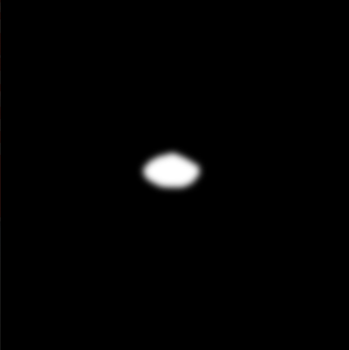}
		\caption{Face \& Mouth}
	\end{subfigure}
	\begin{subfigure}[b]{0.32\columnwidth}
		\includegraphics[width=\textwidth]{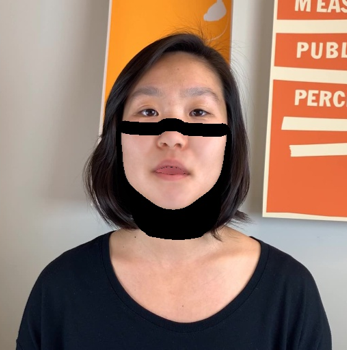}
		\caption{Synth. Comp. ($\mathbf{r}_i$)}
	\end{subfigure}
	\vspace{-0.2cm}
	\caption{
		Training Corpus:
		For each ground truth frame $\mathbf{f}_i$ (a), we obtain a 3D face reconstruction.
		The reconstructed geometry proxy is used to mask out the lower face region (b, left) and render a mouth mask $\mathbf{m}_i$ (b, right), which is used in our training reconstruction loss.
		We superimpose the lower face region from the parametric face model to obtain a synthetic composite $\mathbf{r}_i$ (c).
		The goal of our expression-guided neural renderer is to learn a mapping from the synthetic composite $\mathbf{r}_i$ back to the ground truth frame $\mathbf{f}_i$.
	}
	 \label{fig:data}
\end{figure}

\subsection{Neural Face Rendering}
\label{subsec:rendering}
We employ a novel neural face rendering approach for synthesizing photo-realistic 
talking-head video that matches the modified parameter sequence (\Cref{subsec:param_blend}).
The output of the previous processing step is an edited parameter sequence that describes the new desired facial motion and a corresponding retimed background video clip.
The goal of this synthesis step is to change the facial motion of the retimed background video to match the parameter sequence.
To this end, we first mask out the lower face region, including parts of the neck (for the mask see \Cref{fig:data}b), in the retimed background video and render a new synthetic lower face with the desired facial expression on top.
This results in a video of \emph{composites} $\mathbf{r}_i$ (\Cref{fig:data}d).
Finally, we bridge the domain gap between $\mathbf{r}_i$ and real video footage of the person using our neural face rendering approach, which is based on recent advances in learning-based image-to-image translation \cite{IsolaZZE2017,sun2018hybrid}.

\begin{figure}[t]
	\includegraphics[width=\columnwidth]{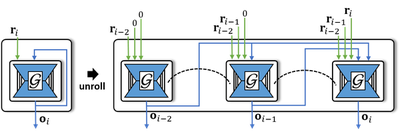}\\
	\vspace{-0.5cm}
	\caption
	{
		We assume the video has been generated by a sequential process, which we model by a recurrent network with shared generator $\mathcal{G}$. In practice, we unroll the loop three times.
	}
	\label{fig:network}
\end{figure}

\begin{figure}[t]
	\includegraphics[width=\columnwidth]{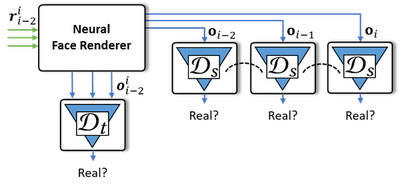}\\
	\vspace{-0.5cm}
	\caption
	{
		We employ a spatial discriminator $\mathcal{D}_s$, a temporal discriminator $\mathcal{D}_t$, and an adversarial patch-based discriminator loss to train our neural face rendering network.
	}
	\label{fig:network2}
\end{figure}

\subsubsection{Training the Neural Face Renderer}
To train our neural face rendering approach to bridge the domain gap we start from a paired training corpus $\mathcal{T} = \big\{(\mathbf{f}_i, \mathbf{r}_i) \big\}_{i=1}^{N}$ that consists of the $N$ original video frames $\mathbf{f}_i$ and corresponding synthetic composites $\mathbf{r}_i$.
The $\mathbf{r}_i$ are generated as described in the last paragraph, but using the ground truth tracking information of the corresponding frame (\Cref{fig:data}), instead of the edited sequence, to render the lower face region.
The goal is to learn a temporally stable video-to-video mapping (from $\mathbf{r}_i$ to $\mathbf{f}_i$) using a recurrent neural network (RNN) that is trained in an adversarial manner.
We train one person-specific network per input video.
Inspired by the video-to-video synthesis work of Wang et al.\,\shortcite{wang2018vid2vid}, our approach assumes that the video frames have been generated by a sequential process, i.e., the generation of a video frame depends only on the history of previous frames (\Cref{fig:network}).
In practice, we use a temporal history of size $L=2$ in all experiments, so the face rendering RNN looks at $L+1=3$ frames at the same time.
The best face renderer $\mathcal{G}^*$ is found by solving the following optimization problem:
\begin{equation}
\mathcal{G}^*
=
\argmin_{\mathcal{G}}
{
	\max_{\mathcal{D}_s,\mathcal{D}_t}
	{
		\mathcal{L}(\mathcal{G}, \mathcal{D}_s, \mathcal{D}_t)
	} 
}
\enspace{.}
\end{equation}
Here, $\mathcal{D}_s$ is a per-frame spatial patch-based discriminator \cite{IsolaZZE2017}, and $\mathcal{D}_t$ is a temporal patch-based discriminator.
We train the recurrent generator and the spatial and temporal discriminator of our GAN in an adversarial manner, see \Cref{fig:network2}.
In the following, we describe our training objective $\mathcal{L}$ and the network components in more detail.

\paragraph{Training Objective}
For training our recurrent neural face rendering network, we employ stochastic gradient decent to optimize the following training objective:
\begin{equation}
\mathcal{L}(\mathcal{G}, \mathcal{D}_s, \mathcal{D}_t)
=
\mathbb{E}_{(\mathbf{f_i}, \mathbf{r_i})}
\big[\mathcal{L}_{r}(\mathcal{G}) + \lambda_s \mathcal{L}_{s}(\mathcal{G}, \mathcal{D}_s) + \lambda_t \mathcal{L}_{t}(\mathcal{G}, \mathcal{D}_t)\big]
\enspace{.}
\end{equation}
Here, $\mathcal{L}_{r}$ is a photometric reconstruction loss, $\mathcal{L}_{s}$ is a per-frame spatial adversarial loss, and $\mathcal{L}_{t}$ is our novel adversarial temporal consistency loss that is based on difference images.
Let $\mathbf{f}_{i-L}^i$ denote the tensor of video frames from frame $\mathbf{f}_{i-L}$ to the current frame $\mathbf{f}_i$.
The corresponding tensor of synthetic composites $\mathbf{r}_{i-L}^i$ is defined in a similar way.
For each of the $L+1$ time steps, we employ an $\ell_1$-loss to enforce the photometric reconstruction of the ground truth:
\begin{eqnarray}
\mathcal{L}_{r}(\mathcal{G}) =
\sum_{l=0}^{L} \Big|\Big| m_{i-L+l} \otimes \big(\mathbf{f}_{i-L+l} - \mathcal{G}(\mathbf{c}_{i,l})\big) \Big|\Big|_1
 \enspace{,} \nonumber \\
 \text{with~}~\mathbf{c}_{i,l} = \big( \mathbf{r}_{i-L}^{i-L+l}, \mathbf{o}_{i-L}^{i-L+l-1} \big) \enspace{.}
\end{eqnarray}
Here, the $\mathbf{c}_{i,l}$ are the generator inputs for the current frame $i$ and time step $l$, with $\mathbf{o}_{i-L}^{i-L+l-1}$ being the tensor of output frames for the previous time steps.
$\otimes$ is the Hadamard product and $m_{i-L+l}$ 
is a mouth re-weighting mask that gives a higher weight to photometric errors in the mouth region (\Cref{fig:data}).
The mask is $1$ away from the mouth, $10$ for the mouth region, and has a smooth transition in between.
Note the same generator $\mathcal{G}$ is shared across all time steps.
For each time step, missing outputs of non existent previous frames (we only unroll 3 steps) and network inputs that are in the future are replaced by zeros (\Cref{fig:network}).
In addition to the reconstruction loss, we also enforce a separate patch-based adversarial loss for each frame:
\begin{align}
\mathcal{L}_{s}(\mathcal{G},\mathcal{D}_s)
= \sum_{l=0}^{L} \Big[
\log(&\mathcal{D}_s(\mathbf{r}_{i-L+l}, \mathbf{f}_{i-L+l})) \nonumber \\ 
 &+ \log(1 - \mathcal{D}_s(\mathbf{r}_{i-L+l}, \mathcal{G}(\mathbf{c}_{i,l}))) \Big]
\enspace{.}
\end{align}
Note there exists only one discriminator network $\mathcal{D}_s$, which is shared across all time steps.
We also employ an adversarial temporal consistency loss based on difference images \cite{Martin-Brualla:2018}:
\begin{equation}
\mathcal{L}_{t}(\mathcal{G},\mathcal{D}_t)
=
\log(\mathcal{D}_t(r_{i-L}^{i}, \mathbf{\Delta}_{i,l}(\mathbf{f})))
+
\log(1 - \mathcal{D}_t(r_{i-L}^{i}, \mathbf{\Delta}_{i,l}(\mathbf{o})))
\enspace{.}
\end{equation}
Here, $\mathbf{\Delta}_{i,l}(\mathbf{f})$ is the ground truth tensor and $\mathbf{\Delta}_{i,l}(\mathbf{o})$ the tensor of synthesized difference images.
The operator $\mathbf{\Delta}(\bullet)$ takes the difference of subsequent frames in the sequence:
\begin{equation}
\mathbf{\Delta}_{i,l}(\mathbf{x}) = \mathbf{x}_{i-L+1}^{i} - \mathbf{x}_{i-L}^{i-1} \enspace{.}
\end{equation}

\paragraph{Network Architecture}
For the neural face rendering network, we employ an encoder-decoder network with skip connections that is based on U-Net \cite{RonneFB2015}.
Our spatial and temporal discriminators are inspired by \citet{IsolaZZE2017} and \citet{wang2018vid2vid}.
Our network has 75 million trainable parameters.
\newstuff{All subnetworks ($\mathcal{G}$, $\mathcal{D}_s$, $\mathcal{D}_t$) are trained from scratch, i.e., starting from random initialization.
We alternate between the minimization to train $\mathcal{G}$ and the maximization to train $\mathcal{D}_s$ as well as $\mathcal{D}_t$.
In each iteration step, we perform both the minimization as well as the maximization on the same data, i.e., the gradients with respect to the generator and discriminators are computed on the same batch of images.
We do not add any additional weighting between the gradients with respect to the generator and discriminators as done in \citet{IsolaZZE2017}.
The rest of the training procedure follows \citet{IsolaZZE2017}.
}
For more architecture details, see Supplemental W13.

The rendering procedure produces photo-realistic video frames of the subject, appearing to speak the new phrase $W$.
These localized edits seamlessly blend into the original video, producing an edited result, all derived from text. %

\paragraph{Adding Audio}
The video produced by our pipeline is mute. To add audio we use audio synthesized either by the built in speech synthesizer in
Mac OS X, or by VoCo\,\cite{jin2017voco}.
An alternative is to obtain an actual recording of the performer's
voice.
In this scenario, we retime the resulting video to match the
recording at the level of \newstuff{phones}.
Unless noted otherwise, all of our synthesis results presented in the
performer's own voice are generated using this latter method.
Note that for move and delete edits we use the performer's voice from
the original video.

\begin{figure*}
	\centering      
	\includegraphics[width=\textwidth]{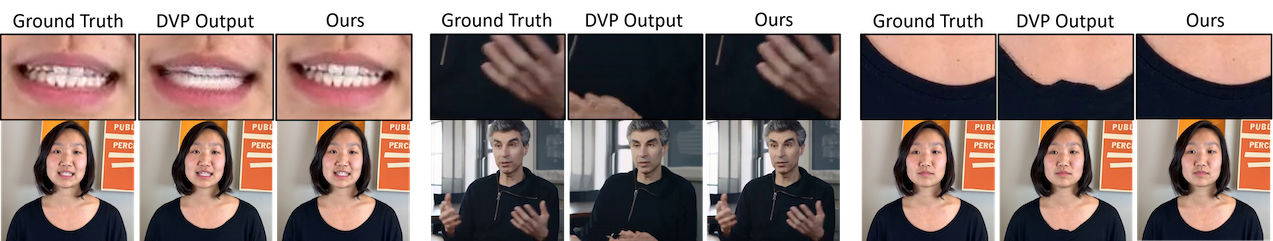}
	\vspace{-0.5cm}
	\caption{
		Comparison of different neural face rendering backends:
		We compare the output of our approach with a baseline that is trained based on input data as proposed in Deep Video Portraits (DVP) \cite{kim2018DeepVideo}.
		DVP does not condition on the background and thus cannot handle dynamic background.
		In addition, this alternative approach fails if parts of the foreground move independently of the head, e.g., the hands.
		Our approach explicitly conditions on the background and can thus handle these challenging cases with ease.
		In addition, our approach only has to spend capacity in the mouth region (we also re-weight the reconstruction loss based on a mouth mask), thus our approach gives much sharper higher quality results.
		\textit{\tiny{\newstuff{Video credit (middle): The Mind of The Universe.}}}
		}
	\label{fig:reenactment}
	\vspace{-0.1cm}
\end{figure*}

\begin{figure*}
	\centering      
	\includegraphics[width=\textwidth]{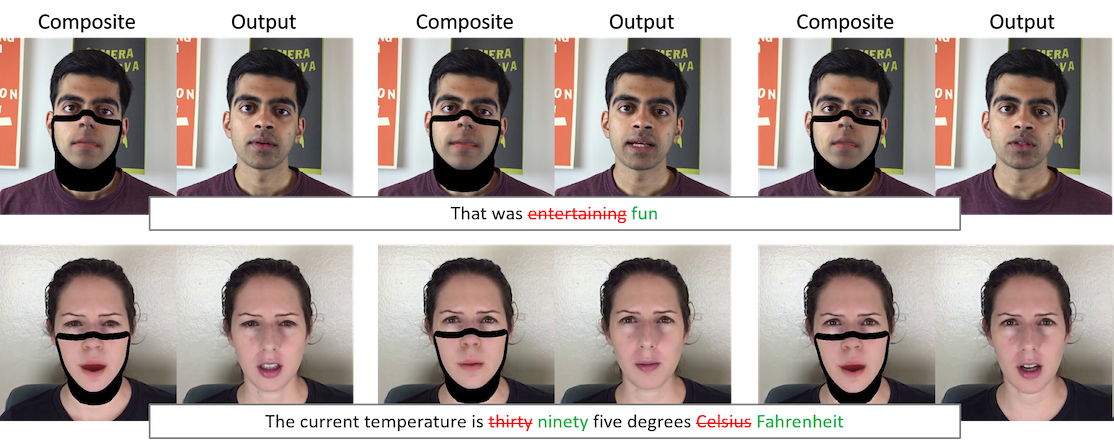}
	\vspace{-0.8cm}
	\caption{
		Our approach enables a large variety of text-based edits, such as deleting, rearranging, and adding new words.
		Here, we show examples of the most challenging of the three scenarios, adding one or more unspoken words.
		As can be seen, our approach obtains high quality reenactments of the new words based on our neural face rendering approach that converts synthetic composites into photo-real imagery.
		For video results we refer to the supplemental.
		}
	\label{fig:figure_edits}
\end{figure*}

\section{Results}
\label{sec:applications}
We show results for our full approach on a variety of videos, both recorded
by ourselves and downloaded from YouTube (\Cref{tbl:inputs}). We encourage the reader to view video results (with audio) in the supplemental video and website,
since our results are hard to evaluate from static frames.

\begin{table}
\label{tbl:inputs}
\begin{center}
\caption{Input sequences. We recorded three sequences, each about 1 hour long. The sequences contain ground truth sentences and test sentences we edit, and also the first 500 sentences from the TIMIT dataset. We also downloaded a 1.5 hour long interview from YouTube that contains camera and hand motion, and an erroneous transcript.
\newstuff{Seq2 and Seq3 are both 60fps. Seq1 was recorded at 240fps, but since our method produces reasonable results with lower frame rates, we discarded frames and effectively used 60fps. Seq4 is 25fps, and still produces good results.
}}
\vspace{-0.15cm}
\begin{tabular}{lccc}
    \toprule
            & Source        & Transcript         & Length   \\
    \midrule
    Seq1    & Our recording & Manually verified             & \textasciitilde1 hour  \\
    Seq2    & Our recording & Manually verified             & \textasciitilde1 hour  \\
    Seq3    & Our recording & Manually verified             & \textasciitilde1 hour  \\
    Seq4    & YouTube       & Automatic (has errors)        & \textasciitilde1.5 hours \\
    \bottomrule
\end{tabular}
\vspace{-0.4cm}
\end{center}
\end{table}

\paragraph{Runtime Performance}
3D face reconstruction takes 110ms per frame.
Phoneme alignment takes 20 minutes for a 1 hour speech video.
Network training takes 42 hours. We train for 600K iteration steps with a batch size of 1.
Viseme search depends on the size of the input video and the new edit. For a 1 hour recording with continuous speech, viseme search 
takes between 10 minutes and 2 hours for all word insertion operations in this paper.
Neural face rendering takes 132ms per frame.
All other steps of our pipeline incur a negligible time penalty.

\subsection{Video Editing}
\begin{figure}
	\centering      
	\includegraphics[width=\columnwidth]{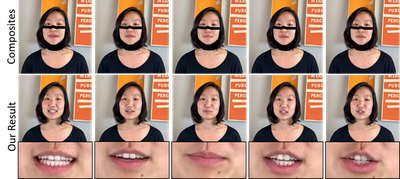}
	\vspace{-0.5cm}
	\caption{
	Our approach synthesizes the non-rigid motion of the lips at high quality (even lip rolling is captured) given only a coarse computer graphics rendering as input.
	In addition, our approach synthesizes a photorealistic mouth interior including highly detailed teeth.
	The synthesis results are temporally coherent, as can be seen in the supplemental video.
	}
	\label{fig:figure_teeth}
\end{figure}

Our main application is text-based editing of talking-head video.
We support moving and deleting phrases, and the more challenging task of adding new unspoken words.
A few examples of replacing one or more words by unspoken word(s) are shown in \Cref{fig:teaser} and \Cref{fig:figure_edits}.
Our approach produces photo-realistic results with good audio to video alignment and a photo-realistic mouth interior including highly detailed teeth (\Cref{fig:figure_teeth}).
For more examples of adding new words, and results for moves and deletes we refer to the supplemental video and Supplemental W1--W4.

\begin{figure}
	\centering      
	\includegraphics[width=\columnwidth]{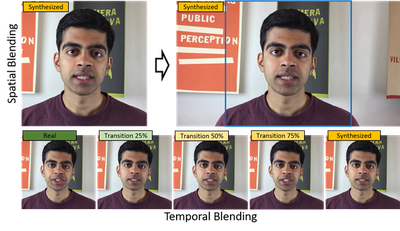}
	\vspace{-0.8cm}
	\caption{
		Our approach enables us to seamlessly compose the modified segments back into the original full frame input video sequence, both spatially as well as temporally.
		We do this by explicitly conditioning video generation on the re-timed background video.
		}
	\label{fig:figure_compositing}
\end{figure}

Our approach enables us to seamlessly re-compose the modified video segments into the original full frame video footage, and to seamlessly blend new segments into the original (longer) video.
Thus our approach can handle arbitrarily framed footage, and is agnostic to the resolution and aspect ratio of the input video. It also enables localized edits (i.e. using less computation) that do not alter most of the original video and can be incorporated into a standard editing pipeline.
Seamless composition is made possible by our neural face rendering strategy that conditions video generation on the original background video.
This approach allows us to accurately reproduce the body motion and scene background (\Cref{fig:figure_compositing}).
Other neural rendering approaches, such as Deep Video Portraits\,\cite{kim2018DeepVideo} do not condition on the background, and thus cannot guarantee that the body is synthesized at the right location in the frame.

\subsection{Translation}
Besides text-based edits, such as adding, rearranging, and deleting words, our approach can also be used for video translation,
\newstuff{as long as the source material contains similar visemes to the target language.}
Our viseme search pipeline is language agnostic. In order to support a new language, we only require a way to convert words into individual phonemes, which is already available for many languages.
We show results in which an English speaker appears to speak German (Supplemental W5).

\subsection{Full Sentence Synthesis Using Synthetic Voice}

With the rise of voice assistants like Alexa, Siri and the Google Assistant, consumers have been getting comfortable with voice-based interactions. We can use our approach to deliver corresponding video. Given a video recording of an actor who wishes to serve as the face of the assistant, our tool could be used to produce the video for any utterance such an assistant might make. We show results of full sentence synthesis using the native Mac OS voice synthesizer (Supplemental W7).
Our system could also be used to easily create instruction videos with more fine-grained content adaptation for different target audiences, or to create variants of storytelling videos that are tailored to specific age groups.

\section{Evaluation, Analysis \& Comparisons}
\label{sec:eval}

To evaluate our approach we have analyzed the content and size of the
input video data needed to produce good results and we have compared
our approach to alternative talking-head video synthesis techniques.

\begin{figure}
	\centering      
	\includegraphics[width=\columnwidth]{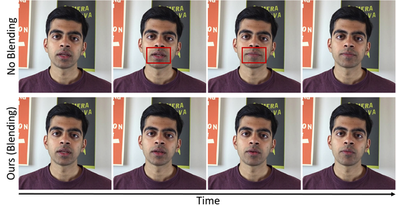}
	\vspace{-0.6cm}
	\caption{Evaluation of Parameter Blending: Without our parameter blending strategy, the editing results are temporally unstable. In this example, the mouth unnaturally closes instantly between two frames without blending, while it closes smoothly with our blending approach.}
	\label{fig:blending2}
\end{figure}

\subsection{Size of Input Video}
\label{subsec:input_size}
We performed a qualitative study on the amount of data required for phoneme retrieval. %
To this end, we iteratively reduced the size of the used training video.
We tested our retrieval approach with 5\%, 10\%, 50\%, and 100\% of the training data (Supplemental W8). 
More data leads in general to better performance and visually more pleasing results, but the quality of the results degrade gracefully with the amount of used data.
Best results are obtained with the full dataset. 

\newstuff{We also evaluate the amount of training data required for our neural face renderer. 
Using seq4 (our most challenging sequence), we test a self-reenactment scenario in which we compare the input frames to our result with varying training data size.  
We obtain errors (mean RMSE per-image) of 0.018 using 100\%, 0.019 using 50\% and 0.021 using only 5\% of the data \newstuff{(R,G,B $\in[0,1]$)}. This result suggests that our neural renderer requires less data than our viseme retrieval pipeline, allowing us to perform certain edits (e.g., deletion) on shorter videos.
}

\subsection{Size of Edit}
We tested our system with various synthesized phrases. We randomly select from a list of ``things that smell'' and synthesize the phrases into the sentence ``I love the smell of X in the morning'' (Supplemental W11). We found that phrase length does not directly correlate with result quality. Other factors, such as the visemes that comprise the phrase and phoneme alignment quality influence the final result.

\subsection{Evaluation of Parameter Space Blending}
We evaluate the necessity of our parameter blending strategy by comparing our approach to a version without the parameter blending (\Cref{fig:blending2} and Supplemental W12).
Without our parameter space blending strategy the results are
temporally unstable.

\subsection{Comparison to MorphCut}
\begin{figure}
	\centering      
	\includegraphics[width=\columnwidth]{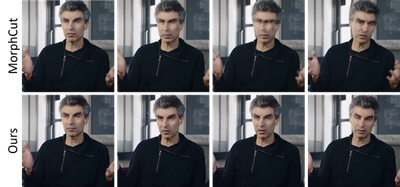}
	\vspace{-0.5cm}
	\caption{
		We compare our approach in the word deletion scenario to \textit{MorphCut}.
		MorphCut fails on the second, third, and forth frames shown here while our approach is able to successfully remove the jump cut.
		\textit{\tiny{\newstuff{Video credit: The Mind of The Universe.}}}
		}
	\label{fig:morph}
\end{figure}
\begin{figure}
	\centering      
	\includegraphics[width=\columnwidth]{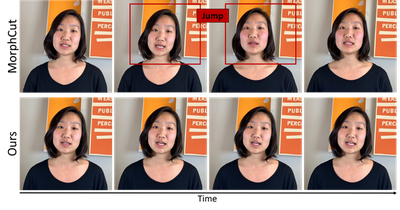}
	\vspace{-0.8cm}
	\caption{
		We tried to stitch retrieved \newstuff{viseme} sequences with MorphCut to generate a new word.
		While our approach with the parameter space blending strategy is able to generate a seamless transition, MorphCut produces a big jump of the head between the two frames.
		}
	\label{fig:morph2}
\end{figure}
\textit{MorphCut} is a tool in Adobe Premiere Pro that is designed
to remove jump cuts in talking-head videos, such as those introduced by moving or deleting words.
It is based on the approach of Berthouzoz et al.\,\shortcite{Berthouzoz:2012:TPC:2185520.2185563}, requires the performer to be relatively still in the video and cannot synthesize new words.
In \Cref{fig:morph}, we compare our approach to MorphCut in the
word deletion scenario and find that our approach is able to
successfully remove the jump cut, while MorphCut fails due to the motion of the head.

We also tried to apply MorphCut to the problem of word addition.
To this end, we first applied our phoneme/viseme retrieval pipeline to select suitable frames to compose a new word.
Afterwards, we tried to remove the jump cuts between the different phoneme subsequences with MorphCut (\Cref{fig:morph2}).
While our approach with parameter space blending is able to generate seamless transitions, MorphCut produces big jumps and can not smooth them out.

\subsection{Comparison to Facial Reenactment Techniques}
\begin{figure}
	\centering      
	\includegraphics[width=\columnwidth]{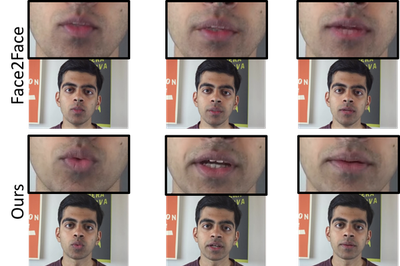}
	\vspace{-0.5cm}
	\caption{
		Comparison to the Face2Face \cite{ThiesZSTN2016a} facial reenactment approach.
		Our approach produces high quality results, while the retrieval-based Face2Face approach exhibits ghosting artifacts and is temporally unstable.
		We refer to the supplemental video for more results.
	}
	\label{fig:reenactment2}
\end{figure}
We compare our facial reenactment backend with a baseline approach that is trained based on the input data as proposed in Deep Video Portraits \cite{kim2018DeepVideo} (\Cref{fig:reenactment}).
For a fair comparison, we trained our recurrent generator network (including the temporal GAN loss, but without our mouth re-weighting mask) with Deep Video Portraits style input data (diffuse rendering, uv-map, and eye conditioning) and try to regress a realistic output video.
Compared to Deep Video Portraits \cite{kim2018DeepVideo}, our approach synthesizes a more detailed mouth region, handles dynamic foregrounds well, such as for example moving hands and arms, and can better handle dynamic background.
We attribute this to our mouth re-weighting mask and explicitly conditioning on the original background and body, which simplifies the learning task, and frees up capacity in the network.
Deep Video Portraits struggles with any form of motion that is not directly correlated to the head, since the head motion is the only input in their technique.
We refer to the supplemental video for more results.

We also compare our approach to Face2Face \cite{ThiesZSTN2016a}, see \Cref{fig:reenactment2}.
Our neural face rendering approach can better handle the complex articulated motion of lips, e.g., lip rolling, and synthesizes a more realistic mouth interior.
The Face2Face results show ghosting artifacts and are temporally unstable, while our approach produces temporally coherent output.
We refer to the supplemental video for more results.

\begin{figure}
	\centering      
	\includegraphics[width=\columnwidth]{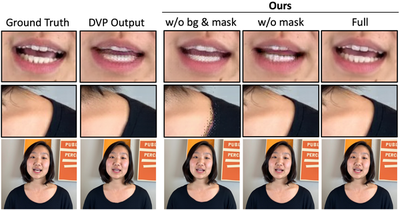}
	\vspace{-0.5cm}
	\caption{
		\newstuff{
		Ablation study comparing ground truth with several versions of our approach:
		a simplified version without providing the mouth mask and the background conditioning (\woBGmask); a simplified version that provides the background but not the mouth mask (\woMask); and our complete approach with all new components (Full).
		In addition, we show a result from the Deep Video Portraits (DVP) of Kim et al.~\shortcite{kim2018deep}.
        All components of our approach positively contribute to the quality of the results, and our full method outperforms DVP.
        This is especially noticeable in the hair and mouth regions. 
		}
	}
	\vspace{-0.4cm}
	\label{fig:ablation}
\end{figure}

\newstuff{
\subsection{Ablation Study}
We also performed an ablation study to evaluate the new components of our approach (see \Cref{fig:ablation}).
We perform the study in a self-reenactment scenario in which we compare our result to the input frames.
To this end we compare our complete approach (Full) with two simplified approaches. The first simplification removes both the mouth mask and background conditioning (w\textbackslash o~bg~\&~mask) from our complete approach, while the second simplification only removes the mouth mask (w\textbackslash o mask).
As shown in \Cref{fig:ablation}, all components positively contribute to the quality of the results.
This is especially noticeable in the mouth region, where the quality and level of detail of the teeth is drastically improved.
In addition, we also show the result obtained with the Deep Video Portraits (DVP) of Kim et al.~\shortcite{kim2018deep}.
We do not investigate alternatives to the RNN in our ablation study, as Wang et al.~\shortcite{wang2018vid2vid} have already demonstrated that RNNs outperform independent per-frame synthesis networks.
}

\subsection{User Study}
\begin{table*}\small
	\renewcommand{\tabcolsep}{0.17cm}
	\begin{center}
	\caption{\label{tab:study}
		We performed a user study with $N=138$ participants and collected in total $2993$ responses to evaluate the quality of our approach.
		Participants were asked to respond to the statement ``This video clip looks real to me'' on a 5-point Likert scale from 1 (\textit{strongly disagree}) to 5 (\textit{strongly agree}).
		We give the percentage for each score, the average score, and the percentage of cases the video was rated as `real' (a score of 4 or higher).
		\newstuff{The difference between conditions is statistically significant (Kruskal-Wallis test, $p < 10^{-30}$). 
		Our results are different from both GT-base and from GT-target (Tukey's honest significant difference procedure, $p < 10^{-9}$ for both tests).
		This suggests that while our results are often rated as real, they are still not on par with real video.}
	}
	\vspace{-0.15cm}
	\begin{tabular}{cccccccccccccccccccccc}
		\toprule
		& \multicolumn{7}{c}{\textbf{GT Base Videos}} & \multicolumn{7}{c}{\textbf{GT Target Videos}} & \multicolumn{7}{c}{\textbf{Our Modified Videos}} \\ 
		\cmidrule(lr){2-8} \cmidrule(lr){9-15} \cmidrule(lr){16-22}
		& \multicolumn{6}{c}{Scores} &  & \multicolumn{6}{c}{Scores} &  & \multicolumn{6}{c}{Scores} &    \\ 
		\cmidrule(lr){2-6} \cmidrule(lr){9-13} \cmidrule(lr){16-20}
		& 5 &  4 &  3 &  2 &  1 & $\Sigma$ & `real'   &  5 &  4 &  3 &  2 &  1 & $\Sigma$ & `real' &  5 &  4 &  3 &  2 &  1 & $\Sigma$ & `real' \\ 
		\cmidrule(lr){2-2}\cmidrule(lr){3-3}\cmidrule(lr){4-4}\cmidrule(lr){5-5}\cmidrule(lr){6-6}\cmidrule(lr){7-7}\cmidrule(lr){8-8}\cmidrule(lr){9-9}\cmidrule(lr){10-10}\cmidrule(lr){11-11}\cmidrule(lr){12-12}\cmidrule(lr){13-13}\cmidrule(lr){14-14}\cmidrule(lr){15-15}\cmidrule(lr){16-16}\cmidrule(lr){17-17}\cmidrule(lr){18-18}\cmidrule(lr){19-19}\cmidrule(lr){20-20}\cmidrule(lr){21-21}\cmidrule(lr){22-22}

		\textbf{Set~1}     & 45.3 & 36.3 & 7.9 & 10.0 & 0.5 & 4.1 &    81.6\% & 47.0 & 31.9 & 9.7 & 10.1 & 1.4 & 4.1 &  78.9\%    & 31.9 & 25.2 & 10.9 & 23.9 & 8.2 & 3.5 &   57.1\% \\ %
		\textbf{Set~2}     & 41.6 & 38.1 & 9.9 &  9.2 & 1.2 & 4.1 &    79.7\% & 45.7 & 39.8 & 8.7 &  5.4 & 0.4 & 4.3 &  85.6\%    & 29.3 & 32.8 &  9.4 & 22.9 & 5.7 & 3.9 &   62.1\% \\ %
		\midrule   
		\textbf{Mean}      & 43.5 & 37.2 & 8.9 &  9.6 & 0.9 & 4.1 &    80.6\% & 46.4 & 35.9 & 9.2 &  7.7 & 0.9 & 4.2 &  82.2\%    & 30.6 & 29.0 & 10.1 & 23.4 & 7.0 & 3.7 &   59.6\% \\

		\bottomrule
	\end{tabular}
	\end{center}
\end{table*}

To quantitatively evaluate the quality of videos generated by our text-based editing system, we performed a web-based user study with $N=138$ participants and collected $2993$ individual responses, see \Cref{tab:study}.
The study includes videos of two different talking heads,
\textit{Set~1} and \textit{Set~2}, where each set contains 6 different base sentences.
For each of the base sentences, we recorded a corresponding target sentence in which one or more words are different.
We use both the base and target sentences as ground truth in our user study.
Next, we employed our pipeline to artificially change the base into the target sentences.
In total, we obtain $2\times 3\times 6=36$ video clips.

In the study, the video clips were shown one video at a time to 
participants $N=138$ in randomized order and they were asked to respond to the statement ``This
video clip looks real to me'' on a 5-point Likert scale
(5-\textit{strongly agree}, 4-\textit{agree}, 3-\textit{neither
	agree nor disagree}, 2-\textit{disagree}, 1-\textit{strongly disagree}).
As shown in \Cref{tab:study}, the real ground truth base videos were only rated to be `real' $80.6\%$ of the cases and the real ground truth target videos were only rated to be `real' $82.2\%$ of the cases (score of 4 or 5).
This shows that the participants were already highly alert, given they were told it was a study on the topic of `Video Editing'.
Our pipeline generated edits were rated to be `real' $59.6\%$ of the cases, which means that 
more than half of the participants found those
clips convincingly real.
\Cref{tab:study} also reports the percentage of times each score was given and the average score per video set.
Given the fact that synthesizing convincing audio/video content is very challenging, since humans are highly tuned to the slightest audio-visual misalignments (especially for faces), this evaluation shows that our approach already achieves compelling results in many cases.

\section{Limitations \& Future Work}
\label{sec:limiations}

While we have demonstrated compelling results in many challenging scenarios, there is room for further improvement and follow-up work:
(1) Our synthesis approach requires a re-timed background video as input.
Re-timing changes the speed of motion, thus eye blinks and gestures might not perfectly align with the speech anymore.
To reduce this effect, we employ a re-timing region that is longer than the actual edit, thus modifying more of the original video footage, with a smaller re-timing factor.
For the insertion of words, this could be tackled by a generative model that is able to synthesize realistic complete frames that also include new body motion and a potentially dynamic background.
(2) Currently our phoneme retrieval is agnostic to the mood in which the phoneme was spoken.
This might for example lead to the combination of happy and sad segments in the blending.
Blending such segments to create a new word can lead to an uncanny result.
(3) Our current viseme search aims for quality but not speed. We would like to explore approximate solutions to the viseme search problem, which we believe can allow interactive edit operations.
\newstuff{
(4) We require about 1 hour of video to produce the best quality results. 
To make our method even more widely applicable, we are investigating ways to produce better results with less data. 
Specifically, we are investigating ways to transfer expression parameters across individuals, which will allow us to use one pre-processed dataset for all editing operations.
}
(5) Occlusions of the lower face region, for example by a moving hand, interfere with our neural face renderer and lead to synthesis artifacts, since the hand can not be reliably re-rendered.
Tackling this would require to also track and synthesize hand motions.
Nevertheless, we believe that we demonstrated a large variety of compelling text-based editing and synthesis results.
In the future, end-to-end learning could be used to learn a direct mapping from text to audio-visual content.

\section{Conclusion}
\label{sec:conclusion}

We presented the first approach that enables text-based editing of talking-head video by modifying the corresponding transcript.
As demonstrated, our approach enables a large variety of edits, such as addition, removal, and alteration of words, as well as convincing language translation and full sentence synthesis.
We believe our approach is a first important step towards the goal of fully text-based editing and synthesis of general audio-visual content.

\begin{acks}
\newstuff{
This work was supported by 
the Brown Institute for Media Innovation,
the Max Planck Center for Visual Computing and Communications, 
ERC Consolidator Grant 4DRepLy (770784),
Adobe Systems,
and the Office of the Dean for Research at Princeton University.
}
\end{acks}

\bibliographystyle{ACM-Reference-Format}
\bibliography{main}

\appendix
\begin{figure}[b]
	\includegraphics[width=\columnwidth]{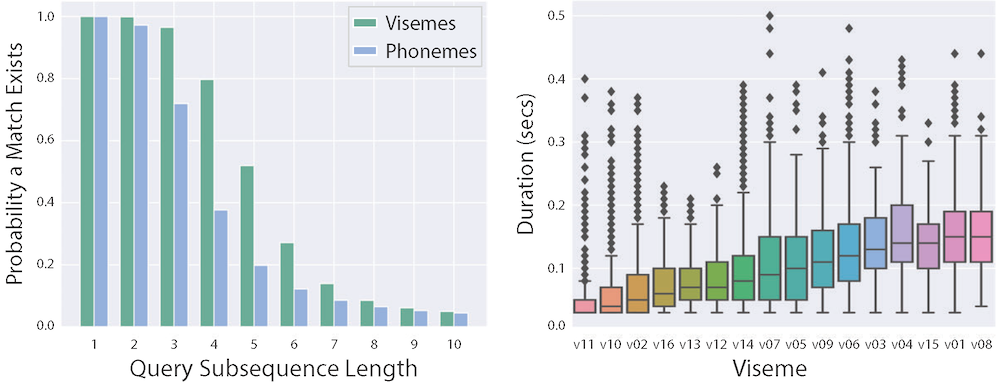}\\
 	\vspace{-0.2cm}
	\caption
	{Left: probability of matching phoneme/viseme subsequences of length $K \in [1, 10]$ in the TIMIT corpus. To ensure that the query subsequences reflect the distribution of such sequences in English we employ a leave-one-out strategy: we choose a random TIMIT sequence of length K, and look for an exact match anywhere in the rest of the dataset. Exact matches of more than 4-6 vismes and 2-3 phonemes are uncommon.
	Right: variation in viseme duration in TIMIT. Different instances of a single viseme vary by up to an order of magnitude. Between different visemes, the median instance length varies by a factor of five.
	}
	\label{fig:match-probability}
\end{figure}

\section{Phoneme \& Viseme Content}
\label{subsec:data_analysis}

Our matching algorithm (\Cref{subsec:viseme_search}) is
designed to find the longest match between subsequences of
phonemes/visemes in the edit and the input video. Suppose our input
video consists of all the sentences in the TIMIT corpus \cite{timit}, 
a set that has been designed to be phonetically rich by
acoustic-phonetic reseachers. 
\Cref{fig:match-probability} plots the probability of finding an exact match anywhere in TIMIT to a phoneme/viseme subsequence of length $K \in [1, 10]$.
Exact matches of more than 4-6 visemes or 3-5 phonemes are rare. This
result suggests that even with phonetically rich input video we cannot
expect to find edits consisting of long sequences of phonemes/visemes
(e.g. multiword insertions) in the input video and that our approach
of combining shorter subsequences with parameter blending is
necessary.

\Cref{fig:match-probability} also examines the variation in individual
viseme instances across the set of 2388 sentences in
the TIMIT corpus. We see that there is variation both between
different visemes and within a class of visemes. These observations
led us to incorporate viseme distance and length in our search
procedure (\Cref{subsec:viseme_search}) and informed our blending
strategy (\Cref{subsec:param_blend}).

\end{document}